\documentclass[a4paper]{article}
\usepackage[latin1]{inputenc} % ou \usepackage[utf8]{inputenc}
\usepackage[T1]{fontenc} % ou \usepackage[OT1]{fontenc}
\usepackage{RR,RRthemes}
\usepackage[colorlinks=true,linkcolor=blue]{hyperref}
\usepackage{amsmath}
\usepackage{amssymb}
\usepackage{xcolor}
\usepackage{graphicx}
\usepackage{flafter}  % floats come never before where they are in the text
\RRdate{October 2011}
\RRNo{7748}
\RRauthor{% les auteurs
Nikolaus Hansen%\thanks{INRIA}%
}
\authorhead{N.\ Hansen}
\RRtitle{} % french title
\RRetitle{Injecting External Solutions Into CMA-ES}
\titlehead{Injecting External Solutions Into CMA-ES}
\RRnote{}
\RRnote{}
\RRabstract{This report considers how to inject external candidate solutions into the CMA-ES algorithm. The injected solutions might stem from a gradient or a Newton step, a surrogate model optimizer or any other oracle or search mechanism. They can also be the result of a repair mechanism, for example to render infeasible solutions feasible. Only small modifications to the CMA-ES are necessary to turn injection into a reliable and effective method: too long steps need to be tightly renormalized. The main objective of this report is to reveal this simple mechanism. 

Depending on the source of the injected solutions, interesting variants of CMA-ES arise. When the best-ever solution is always (re-)injected, an elitist variant of CMA-ES with weighted multi-recombination arises. When \emph{all} solutions are injected from an \emph{external} source, the resulting algorithm might be viewed as \emph{adaptive encoding} with step-size control. 

In first experiments, injected solutions of very good quality lead to a convergence speed twice as fast as on the (simple) sphere function without injection. This means that we observe an impressive speed-up on otherwise difficult to solve functions. Single bad injected solutions on the other hand do no significant harm. 
}
\RRkeyword{CMA-ES, external solutions, gradient, injection, repair}
\RRprojet{TAO}  % cas d'un seul projet
\RRdomaine{1} % cas du domaine numero 1
\RRthemeProj{tao} % theme du projet Apics
\RRdomaineProjBis{tao} % domaine du projet Apics
\RCSaclay % Saclay \^Ile de France
\newcommand{\oeil}[1]{\textbf{#1}}
\newcommand{\mc}[2]{\newcommand{#1}{\ensuremath{#2}}} % math command without argument
\mc{\rootinvC}{\C^{-1/2}}
\mc{\rootC}{\C^{1/2}}

\newcommand{\ve}[1]{\ensuremath{{\mathbf{#1}}}}
\newcommand{\ma}[1]{\ensuremath{{\mathbf{#1}}}}

\newcommand{\zz}{\vspace{-0.7ex}}
\newcommand{\R}[1][]{\ensuremath{\mathbb{R}^{#1}}} %
\newcommand{\Rn}{\R[\dim]}
\newcommand{\T}{{\mathrm{T}}}
\newcommand{\Ctminushalf}{{\Ct}^{{-\frac{1}{2}}}}
\newcommand{\Cminushalf}{{\C}^{{-\frac{1}{2}}}}
\newcommand{\NormalNI}{\ensuremath{{\Normal{\ve{0},\ve{I}}}}}
\newcommand{\Normal}[1]{\ensuremath{\mathcal{N}\hspace{-0.13em}\left(#1\right)}}

\newcommand{\x}{\ensuremath{\ve{x}}}
\newcommand{\y}{\ensuremath{{\ve{y}}}}
\newcommand{\xmean}{\ve{m}}
\newcommand{\m}{\xmean}
\newcommand{\lam}{\ensuremath{\lambda}}
\newcommand{\sig}{\ensuremath{\sigma}}

\newcommand{\Hsig}{\ensuremath{{h_\sig}}}
\newcommand{\cmu}{\ensuremath{{c_{\mu}}}}

\newcommand{\B}{\ensuremath{\ve{B}}}

\newcommand{\C}{\ensuremath{\ma{C}}}

\newcommand{\cm}{\ensuremath{c_\mathrm{m}}}
\newcommand{\cc}{\ensuremath{c_\mathrm{c}}}
\newcommand{\pc}{\ensuremath{\ve{p}_\mathrm{c}}}

\newcommand{\cs}{c_\sig}
\newcommand{\ds}{d_\sig}
\newcommand{\ps}{\ensuremath{\ve{p}_\sig}}
\newcommand{\g}{\ensuremath{t}}
\newcommand{\mt}{\ensuremath{\m^t}}
\newcommand{\mtt}{\ensuremath{\m^{t+1}}}
\newcommand{\Ct}{{\ensuremath{{\ve{C}^{{}t}}}}}
\newcommand{\Ctt}{\ensuremath{{\ve{C}^{t+1}}}}
\newcommand{\iol}{\ensuremath{{i=1,\dots,\lam}}}

\newcommand{\ilam}{\ensuremath{{i:\lam}}}
\newcommand{\pct}{\ensuremath{\ve{p}_\mathrm{c}^t}}
\newcommand{\pctt}{\ensuremath{{\ve{p}_\mathrm{c}^{t+1}}}}
\newcommand{\pst}{\ensuremath{\ve{p}_\sig^t}}
\newcommand{\pstt}{\ve{p}_\sig^{t+1}}
\newcommand{\chiN}{\ensuremath{\mathsf{E}\|\NormalNI\|}}
\newcommand{\mueff}{\ensuremath{\mu_\mathrm{\textsc{w}}}}
\newcommand{\mmw}{\ensuremath{{\mu/\mu_\mathrm{\textsc{\protect\raisebox{-0.2ex}{w}}}}}}
\newcommand{\mwl}{(\ensuremath{{\mmw},\lambda})}

\newcommand{\sigt}{\ensuremath{{\sigma^t}}}
\newcommand{\sigtt}{\ensuremath{{\sigma^{t+1}}}}

\newcommand{\new}[1]{\colorbox{pink!35!white}{#1}}

\newcommand{\summ}[2]{\sum_{#1=1}^{#2}}
\newcommand{\rklam}[1]{\left(#1\right)}
\renewcommand{\dim}{\ensuremath{{n}}}
\newcommand{\alphacov}{\ensuremath{\alpha_\mathrm{cov}}}
\newcommand{\Dsigmax}{\ensuremath{\Delta_\sigma^\mathrm{max}}}

\begin{document}
\makeRR   % cas d'un rapport de recherche
%% \makeRT % cas d'un rapport technique.
%% a partir d'ici, chacun fait comme il le souhaite

\tableofcontents

\section{Introduction}
The CMA-ES (Covariance Matrix Adaptation Evolution Strategy \cite{Hansen2001,Hansen:2003,hansen2004}) is a search stochastic algorithm for non-convex continuous optimization in a black-box setting, where we want minimize the objective function (or fitness function)
\[
    f:\Rn\to\R,\; \x\mapsto f(\x)
\] 
without exploiting any a priori specified structure of $f$. 
The CMA-ES algorithm entertains a multivariate normal sampling distribution for \x\ and updates the distribution parameters with a comparatively sophisticated procedure, see Figure~\ref{fig:CMA}. While the algorithm is quite robust to large irregularities in the objective function $f$, even small changes of the update procedure can lead to a dramatic break down of its performance. This property has been perceived as a main weakness of the algorithm. 

In this report we show how to make CMA-ES robust to (almost) arbitrary changes of the solutions used in the update procedure. In other words, we reveal the measures to properly 
inject external proposals for either candidate solution points or directions into the CMA-ES algorithm by replacing some of the \emph{internal} solutions originally sampled by CMA-ES, or, equivalently, use solutions that are modified in any desired way (for example to make them feasible). 

External or modified proposal solutions or directions can have a variety of sources. 
\begin{itemize}
\item a gradient or Newton direction;
\item an improved solution, for example the result of a local search started from a solution sampled by CMA-ES (Lamarckian learning), which allows to use CMA-ES in the context of memetic algorithms;
\item a repaired solution, for example from a previously infeasible solution;
\item an optimal solution of a surrogate model built from already evaluated solutions;
\item the best-ever solution seen so far;
\item proposals from any algorithm running in parallel to CMA-ES (migration). 
\end{itemize}
Because injecting a single bad solution essentially corresponds to decreasing the population size by one, no particular care needs to be taken that \emph{only} (exceptionally) good solutions are introduced. Any promising source of solutions might be used. Within CMA-ES, solutions are sampled symmetrically and therefore also virtually never lead to a systematic improvement before selection. 

When all originally sampled internal solutions are replaced, the resulting procedure resembles \emph{adaptive encoding} \cite{hansen2008aeh}. The main differences to adaptive encoding are: (i) external solutions are represented in the original (phenotypic) space (ii) step-size control remains in place and (iii) the parameter setting is different. Using a different (genotyp) representation to \emph{generate} new external solutions is the crucial idea of adaptive encoding and can also be employed here.  

The modifications introduced in CMA-ES are small but will often be decisive. They are outlined in the next section. 

\paragraph{Notations} 
% In the following we assume to optimize the function $f:\Rn\to\R, \x\mapsto f(x)$ in a black-box scenario. 
 Throughout this report, we use for  $\chiN= \sqrt{2}\, \mathrm{\Gamma}(\frac{n+1}{2})
  /\mathrm{\Gamma}(\frac{n}{2})$ the approximation $\sqrt{\dim}\left(1-\frac{1}{4\dim}+\frac{1}{21\dim^2}\right)$. The notation $a\wedge bc + d$ denotes the minimum of $a$ and $bc+d$. 
  
%%%%%%%%%%%%%%%%%%%%%%%%%%%%%%%%%%%%%%%%%%%%%%%%%%%%%%%%%%%%%%%%%%%%%%%%
\section{Injection in the CMA-ES Algorithm}
%%%%%%%%%%%%%%%%%%%%%%%%%%%%%%%%%%%%%%%%%%%%%%%%%%%%%%%%%%%%%%%%%%%%%%%%
\mc{\cy}{c_y}
\mc{\Deltam}{\Delta\m} %{\y_\mathrm{m}}
\mc{\cym}{c_y^\mathrm{m}}
\mc{\xm}{\x_\mathrm{m}}
\newcommand{\alphaclip}[2]{\ensuremath{\alpha_\mathrm{clip}\!\left(#1, #2\right)}}
\newcommand{\textnormalize}[2]{\ensuremath{\alpha_\mathrm{clip}(#1, #2)}}
%%%%%%%%%%%%%%%%%%%%%%%%%%%%%%%%%%%%%%%%%%%%%%%%%%%%%%%%%%%%%%%%%%%%%%%%
\begin{figure*}
%%%%%%%%%%%%%%%%%%%%%%%%%%%%%%%%%%%%%%%%%%%%%%%%%%%%%%%%%%%%%%%%%%%%%%%%
\rule{\linewidth}{1pt}
\begin{eqnarray}
   \x_i &\sim& \mt + \sigt\times \Ct^{1/2}\Normal{\ve0, \ma I} \qquad\text{for \iol}
      \label{eq:sample}
        \\
    \y_i &=& \frac{\x_\ilam - \mt}{\sigt}
        \qquad\text{where $f(\x_{1:\lam})\le\dots\le f(\x_{\mu:\lam}) \le f(\x_{\mu+1:\lam})\dots$}\\
    \y_i &\gets& \text{\new{$\displaystyle\alphaclip{\cy}{\|\Ctminushalf\y_i\|}\times$\!\!}\;}
    \y_i
    \quad \text{if $\x_{i:\lam}$ was injected}
    \label{eq:normalizey}
        \\
    \Deltam &=& \left\{\begin{array}{ll}
    \text{\new{$\displaystyle\frac{\xm-\mt}{\sigt}$}} &\text{if \xm\ was injected}
        \\
    \displaystyle\summ{i}{\mu}w_i\y_i &\text{otherwise}
    \end{array}\right.
    \label{eq:algreco}
        \\
    \mtt &=& \mt + \cm \sigt \Deltam  %\summ{i}{\mu}w_i\y_i
    \label{eq:mean}
        \\[-0ex]
    \Deltam &=& 
        \text{\new{$\displaystyle\alphaclip{\cym}{\sqrt{\mueff}\|\Ctminushalf\Deltam\|}\times$\!\!}\;}
    % \left(1\text{\new{$\displaystyle\wedge
    % \frac{\cym}{\sqrt{\mueff}\|\Ctminushalf\Deltam\|}$}}\right) 
    \Deltam
        \quad\text{if \xm\ was injected or\dots}
    \label{eq:normalizeym}\\
  \pstt &=& (1-\cs)\,\pst %\nonumber \\&&
        + \sqrt{\cs(2-\cs)\mueff}\;
  \Ctminushalf \,\Deltam %\summ{i}{\mu}w_i\y_i  % \frac{\mtt-\mt}{\cm\,\sigt}
%\quad\parbox{0.43\textwidth}{$\Ctminushalf$ is symmetric with positive eigenvalues such that $\Ctminushalf\Ctminushalf$ is the inverse of \Ct. %\footnote%
%{If $\B\La\B^\T=\C$ is an eigendecomposition into an orthogonal matrix \B\ and a diagonal matrix \La, we have $\Cminushalf=\B\La^{-1/2}\B^\T$.} 
%}
\label{eq:algps}\\
%
%
% hsig = sum(self.ps**2) / (1-(1-sp.cs)**(2*self.countiter)) / self.N < 2 + 4./(N+1)
\Hsig&=&\begin{cases} 1 & \text{if~}{\|\pstt\|^2}{} < \dim (1-(1-\cs)^{2(t+1)}) 
  (2 + 4/(\dim+1)) %\left(1.4 + \frac{2}{\dim+1}\right)\chiN
         \label{eq:alghsig} 
  \\ 0 & \text{otherwise}\end{cases}
  %\quad\parbox{0.3\textwidth}{\raggedright stalls the update of \pc\ in \eqref{eq:algpc} when $\sigma$ increases rapidely~}
  \\ 
  \pctt &=& (1-\cc)\,\pct + \Hsig\sqrt{\cc(2-\cc)\mueff} 
     \,\Deltam  %\summ{i}{\mu}w_i\y_i % \,\frac{\mtt-\mt}{\text{\new{\cm\!}}\,\sigt} 
     % \qquad\text{evolution path for the rank-one update of \Ct}
%\quad\text{with~}
%  \Hsig=\left\{
%    \begin{array}{@{}ll}
%      1 &\mbox{if }{\frac{\|\pstt\|}{\sqrt{1-(1-\cs)^{2(t+1)}}} < \left(1.4 + \frac{2}{\dim+1}\right)\chiN}\\
%      0 &\mbox{otherwise}
%    \end{array}\right.
     \label{eq:algpc}     
\\
%  \Ctt &=& % (1 - c_1' - \cmu) \, 
%  \Ct % + (1-\Hsig)\,c_1\cc(2-\cc)) % \ccov)
%%     \nonumber\\&& {} 
%    + c_1\, \big(\underbrace{\pctt{\pctt}^{\,\T}}_{\!\!\!\text{rank one}\!\!\!}
%             {}- \Ct\big)  
%    + \cmu (\underbrace{\Cmu^+}_{\hspace*{-4em}\text{rank $\min(\dim,\mu)$}\hspace*{-4em}}
%            {}- \Ct)
%    \text{\new{\rule[-0.7ex]{0em}{2.8ex}${}-\cminus($}}\!\! 
%        \underbrace{\text{\new{$\Cmu^-$}}}_{\hspace*{-4em}\text{"active"}\hspace*{-4em}}\!\!\!\text{\new{\rule[-0.7ex]{0em}{2.8ex}\!${} - \alphaminusold\Ct - (1-\alphaminusold)\Cmu)$}}
%\label{eqC}\\
  \Ctt &=& (1- c_1' - \cmu ) \, 
  \Ct % + (1-\Hsig)\,c_1\cc(2-\cc)) % \ccov)
%     \nonumber\\&& {} 
    + c_1\underbrace{\pctt{\pctt}^{\,\T}}_{\!\!\!\text{rank one update}\!\!\!}
    {}+ 
    \cmu \summ{i}{\mu} w_i \y_i\y_i^\T
%    (\cmu \text{\,\,\new{\!\!$\!{}+\cminus(1-\alphaminusold)\!$}})\underbrace{\Cmu^+}_{\hspace*{-4em}\text{rank-$\mu$ update}\hspace*{-4em}}
%    \text{\,\,\new{\!\!\rule[-0.7ex]{0em}{2.8ex}$\!{}-\cminus$}}\!\! 
%        \underbrace{\text{\new{$\Cmu^-$}}}_{\hspace*{-4em}\text{"active"}\hspace*{-4em}}
%\!\!\!\text{\new{\rule[-0.7ex]{0em}{2.8ex}\!${} - \alphaminusold\Ct - (1-\alphaminusold)\Cmu)$}}
\label{eqC}\\
   \sigtt &=& \sigt\times
      \exp\rklam{ %\min\rklam{1, 
            \Dsigmax\wedge{}
            \frac{\cs}{\ds}\rklam{\frac{\|\pstt\|}{\chiN}-1}}
    \qquad\parbox{0.45\textwidth}{\raggedright 
         }
       \label{eq:algsig} 
\end{eqnarray}
\rule{\linewidth}{1pt}
 \caption[CMA]{\label{fig:CMA}Update equations for the state variables in the \mwl-CMA-ES with iteration index $t=0,1,2,\dots$ and $\mt\in\Rn$, $\sigt\in\R_+$, $\Ct\in\R[\dim\times \dim]$ positive definite, $\pst,\pct\in\Rn$ and $\ps^{t=0}=\pc^{t=0}=\ve{0}$, $\C^{t=0}=\ma{I}$ and parameters taken from Table~\ref{tabdef}.
 We have additionally $c_1' = c_1 (1 - (1-\Hsig^2) \cc (2-\cc)$. 
The chosen ordering of equations allows to remove the time index. 
The symbol $\x_\ilam$ is the $i$-th best of the solutions $\x_1,\dots,\x_\lam$. 
The ``optimal'' ordering of \eqref{eq:normalizey}, \eqref{eq:algreco}, \eqref{eq:mean} and \eqref{eq:normalizeym} is an open issue. 
}
\end{figure*}
%%%%%%%%%%%%%%%%%%%%%%%%%%%%%%%%%%%%%%%%%%%%%%%%%%%%%%%%%%%%%%%%%%%%%%%%

\mc{\vv}{\ve{v}}
The CMA-ES algorithm that tolerates injected solutions is displayed in Fig.~\ref{fig:CMA}. New parts are highlighted with shaded background.  Injected solutions replace $\x_i$ in \eqref{eq:sample}. The decisive function \alphaclip{.}{.} used in \eqref{eq:normalizey} and \eqref{eq:normalizeym} reads 
\begin{equation}\label{eq:normalize}
    \alphaclip{c}{x} = 1 \wedge \frac{c}{x}
    \;.
\end{equation}
% where $x\wedge y$ is the minimum of $x$ and $y$. 
However, different choices for \alphaclip{.}{.} are possible, or even desirable, and discussed below. 
With parameter setting $\cy=\cym=\Dsigmax=\infty$, the original CMA-ES is recovered (in this case, Equations \eqref{eq:normalizey} and \eqref{eq:normalizeym} are meaningless).  

An injected direction, \vv, is used by setting 
\begin{equation}
  \x_i = \mt + \sigt\frac{\sqrt{\dim}}{\|\rootinvC\vv\|}\vv
  \;.
\end{equation}
If \vv\ represents a gradient direction, using instead
\begin{equation}
\x_i = \mt + \sigt\frac{\sqrt{\dim}}{\|\rootC\vv\|}\C\vv
\end{equation}
seems to suggest itself. Remark that internal perturbations in CMA-ES follow $\rootC\NormalNI$, where \NormalNI\ is isotropic and $\|\NormalNI\|\approx\sqrt{\dim}$.\footnote{The \emph{symmetric} Cholesky factor \rootC\ does not supply a \emph{rotated} coordinate system as desired for adaptive encoding. In this case, we sample using $\B\ma D\NormalNI \sim \rootC\NormalNI$, where $\B\ma D:\Rn\to\Rn$ is the linear decoding. } 

The decisive operation for injected solutions is given in Equation~\eqref{eq:normalizey} of Figure~\ref{fig:CMA}.  Their Mahalanobis distance to the distribution mean
is clipped at $\cy\approx\sqrt{\dim}+2$, preventing artificially long steps to enter the adaptation procedure. Additionally, but in most cases rather irrelevant after clipping the single steps, \Dsigmax\ (see Table~\ref{tabdef}) keeps possible step-size changes below the factor $\exp(0.6)\approx1.82$. Otherwise, the depicted algorithm is not further modified (unless $\cym$ is set $<\infty$). We also use the original internal strategy parameters for CMA-ES which seems particularly reasonable if only a smaller fraction of internal solutions is replaced in \eqref{eq:sample}.  

%
% /////////////////////////////////////////////////////////
\begin{table*}
\newcommand{\alphaminusmin}{\ensuremath{\alpha^-_\mathrm{min}}}%
\caption{\label{tabdef}
  Default parameter values of \mwl-CMA-ES taken from \cite{hansen2010negative}, where by definition $\summ{i}{\mu}|w_i|=1$ and 
$\mueff^{-1} = {\summ{i}{\mu}w_i^2}$ and $a - b \,\wedge\, c + d := \min(a-b,c+d)$. Only population size $\lambda$ is possibly left to the users choice (see also \cite{Auger:2005a})
}\zz%%
\[\begin{array}{rcl}
%\begin{eqnarray*}
\hline
\lam&=&4+\lfloor 3 \ln \dim \rfloor %\quad\text{~population size, number of newly sampled candidate solutions in each iteration~}
\\ 
 \mu&=&\left\lfloor\frac{\lam}{2}\right\rfloor %\quad\text{~parent number, number of candidate solution used to update the distribution parameters~}
 \\
 w_{i} &=&
 \frac{\ln\big(\frac{\lam+1}{2}\big)-\ln i}{\summ{j}{\mu}\left(\ln\big(\frac{\lam+1}{2}\big)-\ln j\right)} %\quad\text{~recombination weights for $i=1,\dots,\mu\le\lam/2$~}
 \\ %
 \cy&=&\text{\new{$\sqrt{\dim} + 2\dim/(\dim+2)$}}
 \\
 \cym&=&\text{\new{$\sqrt{2\dim} + 2\dim/(\dim+2)$}}
 \\
 \cm &=& 1 % \quad\text{learning rate for the mean, sometimes interpreted as rescaled mutation with  $\kappa=\frac{1}{\cm}\ge1$ }
 \\ %
 \cs&=&\frac{\mueff+2}{\dim+\mueff+3} %\quad\text{~cumulation constant for step-size~}
  %TODO 5vs3, $1/\cs$ is the respective time constant  % +3 should become +5 ? 
 \\
 \ds &=& 1 + \cs +
 2\,\max\left(0,\,\sqrt{\frac{\mueff-1}{\dim+1}}-1\right) %\quad\parbox{0.59\textwidth}{step-size damping, is usually close to one. This formula might be replaced by $2\mueff/\lambda + 0.3 + \cs $ in the near future~} 
 \\
\cc&=&\frac{4+0\times\mueff/\dim}{\dim+4+0\times2\mueff/\dim} 
 %\quad\text{~cumulation constant for $\pc$, the $0\times{}$ might be removed in near future~}
\\\rule{0mm}{3.3ex} 
% defopts.CMA.ccov1 = '2 / ((N+1.3)^2+mueff)  % learning rate for rank-one update'; 
% defopts.CMA.ccovmu = '2 * (mueff-2+1/mueff) / ((N+2)^2+mueff) % learning rate for rank-mu update'; 
c_1&=&\frac{\alphacov\min(1,\lambda/6)}{(\dim+1.3)^2+\mueff} %\quad\text{~covariance matrix learning rate for the rank one update using $\pc$~}
\\
\cmu&=& %\min\left(
    1-c_1 %,\, 
    \wedge\, 
    \alphacov\frac{\mueff-2+1/\mueff}{(\dim+2)^2+\alphacov\mueff/2}
    % \right) 
    %\quad\text{covariance matrix learning rate for rank-$\mu$ update~} 
\\
\alphacov &=& 2 \quad\text{could be chosen $<2$, e.g. $\alphacov = 0.5$ for noisy problems}
\\ %
\Dsigmax &=& \text{\new{1.0}} \quad\mbox{or even $0.6$}%\quad\parbox[t]{0.86\textwidth}{might become 1 in near future, which has only a negligible effect under neutral selection, because the term to the right of the $\wedge$ in \eqref{eq:algsig} is approximately $\frac{\cs}{\ds}\Normal{0,1/(2\dim)}$ under neutral selection ~}
%\\ %
\\\hline
\end{array}\]
%\end{eqnarray*}
\end{table*}

\begin{figure}\centering
\includegraphics[width=0.8\textwidth, trim=0 0 0 5mm, clip]{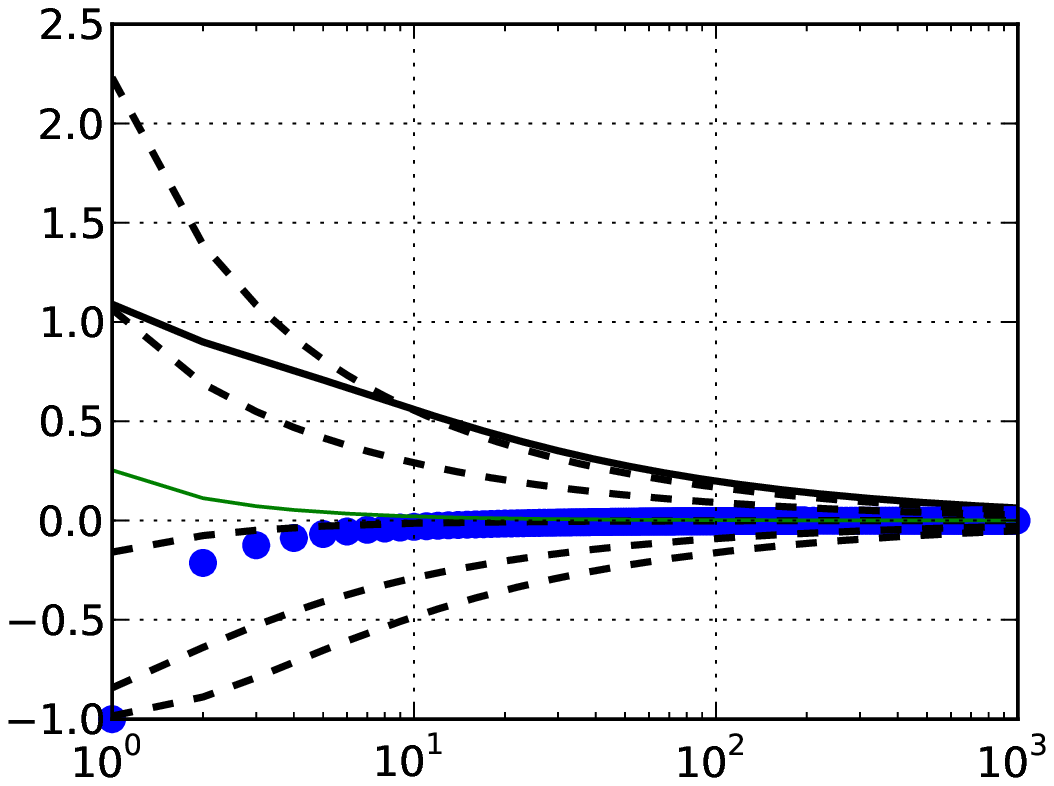}
\includegraphics[width=0.8\textwidth, trim=0 0 0 5mm, clip]{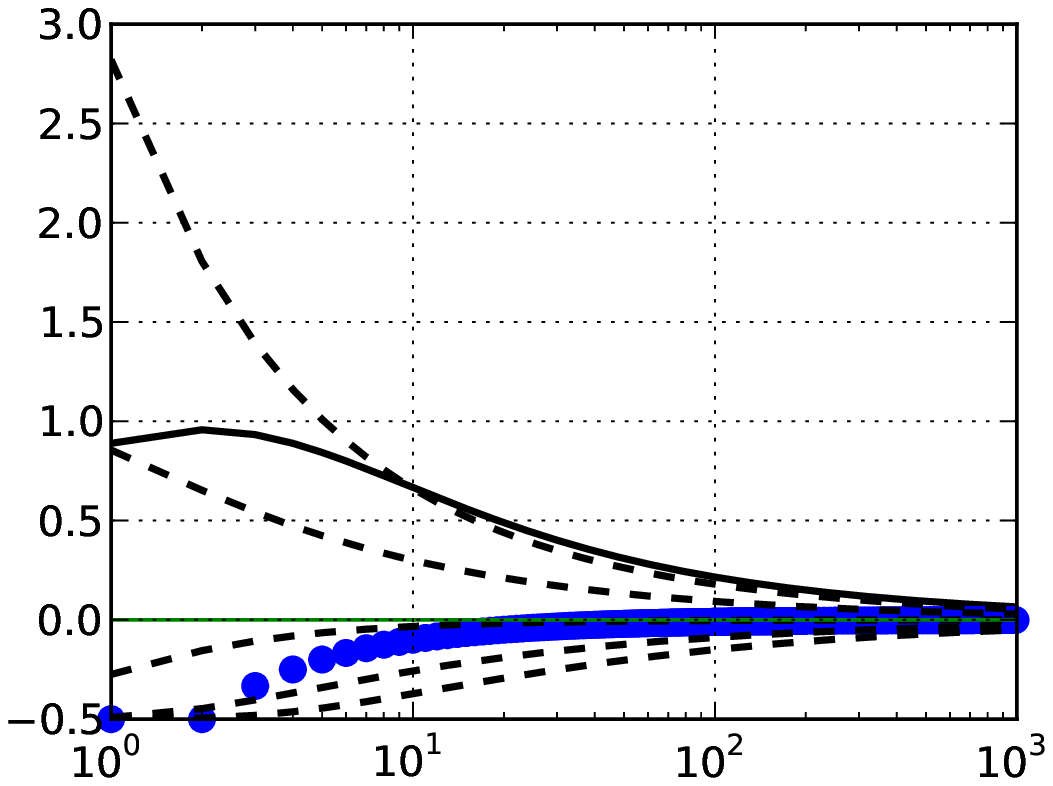}
\caption[]{\label{fig:chilimit}
Relative deviation of $\|\rootinvC\y_i\|$ (top) and $\|\rootinvC\y_i\|^2$ (bottom) from its expected value plotted versus dimension. 
All values are normalized as $x\mapsto x/\chiN - 1$ (top) and $x^2\mapsto (x^2/\dim - 1)/2$ (bottom), compare also \eqref{eq:algsig} for $\cs=\ds=1$. 
Plotted are statistics of the random variable $x$ (top) and $x^2$ (bottom), where $x^2$ follows a chi square distribution with \dim\ degrees of freedom, like $\|\Cminushalf\y_i\|^2$ does without injections under neutral selection. Plotted against \dim\ are modal value ($x=\sqrt{\dim-1}$ and $x^2=0\vee \dim-2$ respectively as dots), the approximation of the expected value $\sqrt{\dim}$ and the expected value $\dim$ respectively (thin solid), the 1, 10, 50, 90, and 99\%tile (dashed) and $x=\cy=\sqrt{\dim} + 2\dim/(\dim+2)$.  }
\end{figure}

\paragraph{Strong injection: mean shift} If we want to make a strong impact with an injection, we can shift the mean. We compute 
\begin{align}
\Deltam &= \frac{\xm-\mt}{\sigt}
\end{align}
from the injected solution \xm\ as in \eqref{eq:algreco}. When no further solutions $\x_i$ are used, the remaining update equations can be performed with $\cmu=0$. With $\cm=1$ (the default), $\mtt=\xm$. In order to prevent an unrealistic large shift of \mt\ in \eqref{eq:mean} we might exchange the order of \eqref{eq:mean} and  \eqref{eq:normalizeym}, therefore applying the length adjustment for \Deltam\ in \eqref{eq:normalizeym} before the actual mean shift \eqref{eq:mean}.  

\paragraph{Parameter setting}
The setting of $\cy\approx\sqrt\dim+2$ is motivated in Figure~\ref{fig:chilimit}. The figure depicts the relative deviation of $\|\rootinvC\y_i\|$ from its expected value. Given its original distribution from CMA-ES, less than 10\% of the $\y_i$ in \eqref{eq:normalizey} are actually clipped. For $\dim>10$, the fraction is smaller than 1\%. 

The typical length of $\sqrt{\mueff}\Deltam$ depends on $\sqrt{\mueff}$ and is often essentially larger than $\sqrt\dim$. Therefore the setting $\cym=\sqrt\dim+2$ leads to a visible impairment of the otherwise unmodified CMA-ES. This suggests that $\cym\approx\sqrt{2\dim}+2$ could be a reasonable choice, however the setting of $\cym$ yet needs further empirical validation. 

In principle, the \oeil{order of Equations} \eqref{eq:normalizey}, \eqref{eq:algreco}, \eqref{eq:mean} and \eqref{eq:normalizeym} can be changed under the constraint that the computation of \Deltam\ in \eqref{eq:algreco} is done before \Deltam\ is used in \eqref{eq:mean} and \eqref{eq:normalizeym}. More specifically, four variants are available, implied by the exchange of \eqref{eq:normalizey} and \eqref{eq:algreco}, or \eqref{eq:mean} and \eqref{eq:normalizeym}, respectively, (another variant that uses unclipped $\y_i$ for \mtt\ but clipped ones for \Deltam\ in the further computations is possible, however not by simple exchange of equations). The variants differ in whether \Deltam\ is computed from clipped $\y_i$ and whether \Deltam\ itself is clipped before or after to compute \mtt. All these variations seem feasible, because an unconstraint shift of \mtt\ is per se not critical for the algorithm behavior. 

%%%%%%%%%%%%%%%%%%%%%%%%%%%%%%%%%%%%%%%%%%%%%%%%%%%%%%%%%%%%%%%%%%%%%%%%%%%%%%%
\section{Discussion}
All update equations starting from \eqref{eq:algreco} are formulated \emph{relative} to the original sample distribution. This means we are, in principle, free to change the distribution before each iteration step. Many reasonable adjustments are possible. A mean shift\footnote{However a mean shift without further updates will impair the meaning of the evolution paths. } (injecting $\xm$ resembles an arbitrary mean shift with additional further updates based on this mean shift), changing the step-size $\sig$, increasing small variances in \C\dots The modification advised in this report is necessary, if $\x_i$ is \emph{not} in accordance with the distribution in \eqref{eq:sample}.  

With the introduced modification(s) the CMA-ES can also be used in the \oeil{adaptive encoding} context (however using $\sig$ for the encoding-decoding might only turn out to be useful if the encoding is an \emph{affine} linear transformation). In the original adaptive encoding \cite{hansen2008aeh}, different normalization measures have been taken for the cumulation in \pc\ and for the covariance matrix update, and the step-size adaptation has been entirely omitted. In this report here, by default the tight normalization of the single steps is the only measure (unless \xm\ is injected). The new normalization replaces the multiplication of the single steps with $\alpha_i$ in \cite{hansen2008aeh} in the covariance matrix update. The new normalization is tighter: choosing $\cy\approx2\sqrt{\dim}$, instead of $\cy\approx\sqrt{\dim}+2$, would be comparable to \cite{hansen2008aeh}. The setting therefore allows to apply step-size control reliably. However, the new setting is less tight for the mean step, as $\cym = \sqrt{\dim}$ (without taking a minimum in \eqref{eq:normalizeym}) would be comparable to \cite{hansen2008aeh}, while we use now $\cym=\infty$ unless an explicite mean-shift is performed. This setting might fail, if all new points point into the same direction viewed from \xm\ (suggesting $\cym \approx 2\sqrt{\dim}$ as a compromise). The new setting seems to be slightly simpler and might turn out preferable also in the adaptive encoding setting, even when leaving aside step-size adaptation. 

Due to the minor modifications we do not expect an adverse interference with \oeil{negative updates} of the covariance matrix as in active CMA-ES \cite{jastrebski2006active,hansen2010negative}. On the contrary, limiting the length of steps that enter the negative update mitigates a principle design flaw of negative updates: long steps tend to be worse (and therefore enter the negative update with a higher probability) and tend to produce a stronger update effect, both just because they are long and not because they indicate an undesirable direction.  

Finally, it is well possible to inject the same solution several times, for example based on its superior quality. One might, for example, consider to unconditionally (re-)inject the best-ever solution in every iteration. Then, an ``elitist algorithm with comma selection'' arises---introducing an easy and appealing way to \oeil{implement elitism} in evolution strategies with weighted multi-recombination.  

A \oeil{generalized approach} to normalize injected solutions compares the empirical CDF of the lengths $l_i^\g=\|\Ctminushalf\y_i\|$, $i=1,\dots,\lambda$, $\g=1,2,\dots$, with a desired CDF, $F$, and reduces the length of $\y_i$ such that the observed relative frequency of lengths larger than or equal to $l_i^\g$ is below, say, $1.2 (1-F(l_i^\g))$. In \eqref{eq:normalize}, the desired CDF is very crudely chosen to be $F(x)=1_{x<\cy}$.  The desired operation in theory is $l_i^\g\gets F_\mathrm{desired}^{-1}(F_\mathrm{true}(l_i^\g))$. A practicable implementation could compare $l_i^\g=\sqrt{2}\times (\|\Ctminushalf\y_i\| - \chiN)$ with the standard normal distribution $\cal F$, in that a correction is applied if $l_i^\g>1$ and $\frac{1}{\g}\frac{1}{\lambda}\sum_{j,k} 1_{l_j^k \ge l_i^\g} > 1.2 (1-{\cal F}(l_i^\g))$. 

\section{Preliminary Experiments}
\begin{figure}
\begin{tabular}{@{\hspace{-3mm}}c@{}c}
 10-D & 40-D
\\
\includegraphics[width=0.499\textwidth, trim=0 0 11mm 8mm, clip]{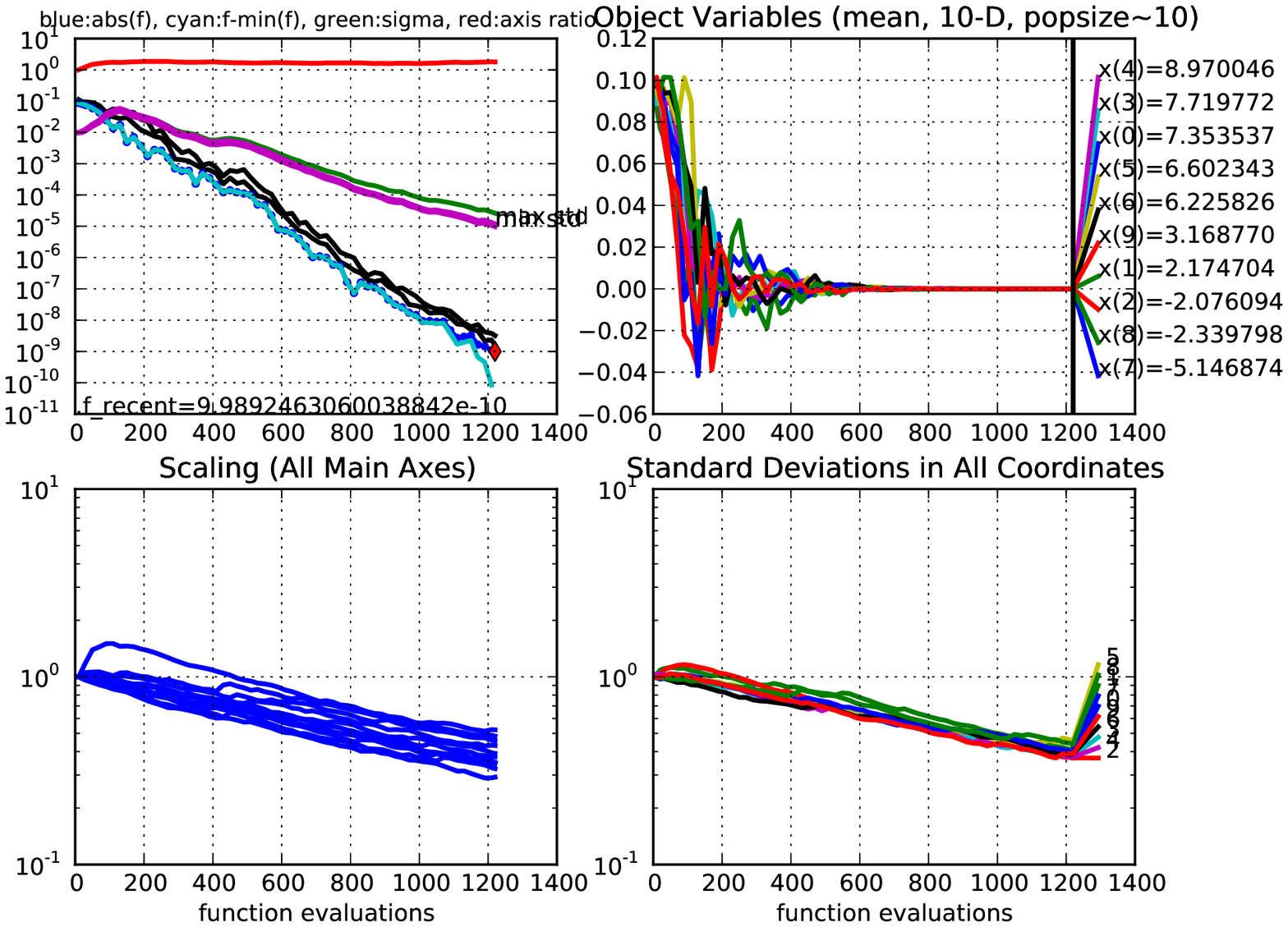}
&\includegraphics[width=0.499\textwidth, trim=0 0 11mm 8mm, clip]{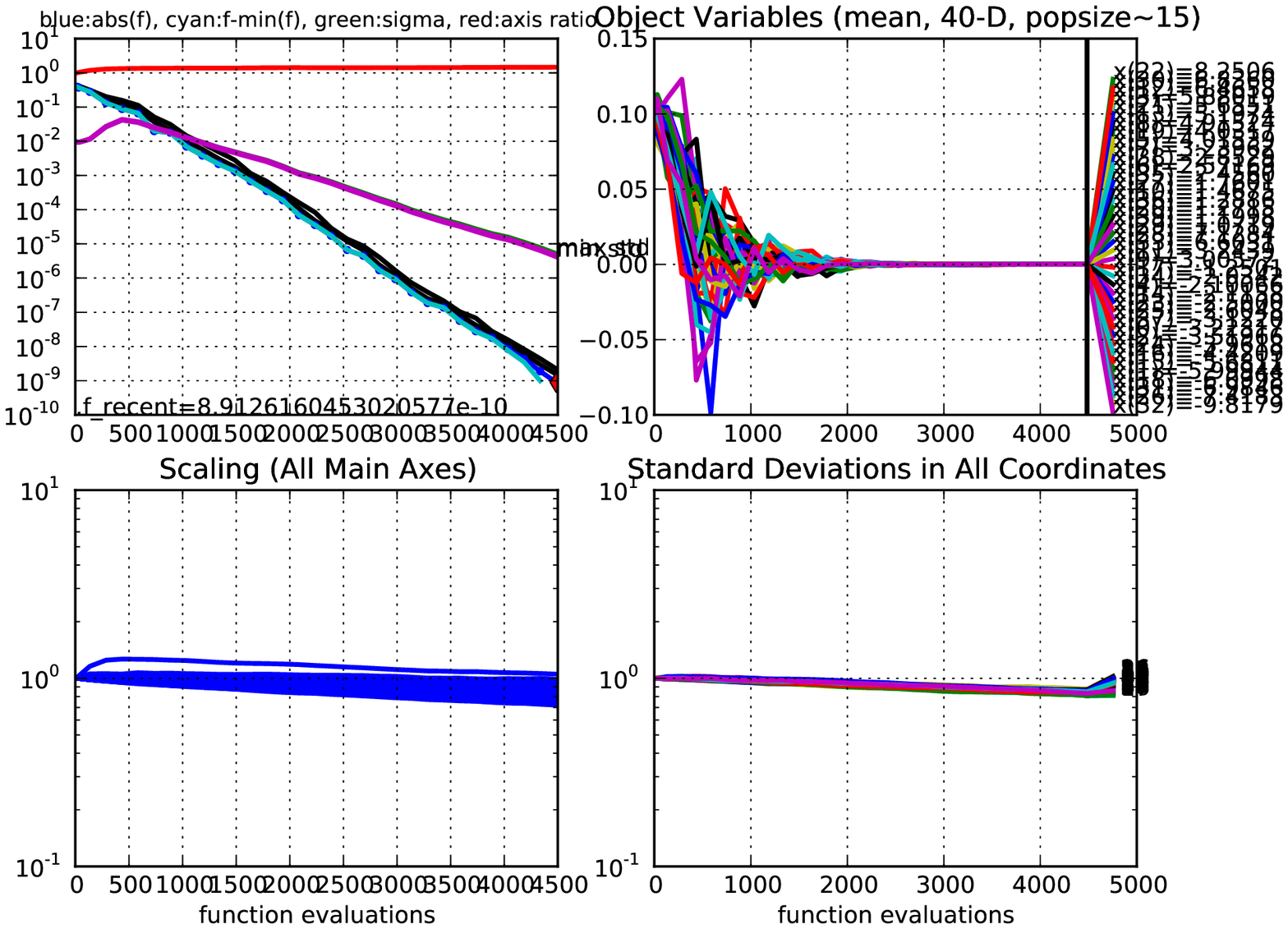}
\\\hline
\includegraphics[width=0.49\textwidth, trim=0 0 11mm 8mm, clip]{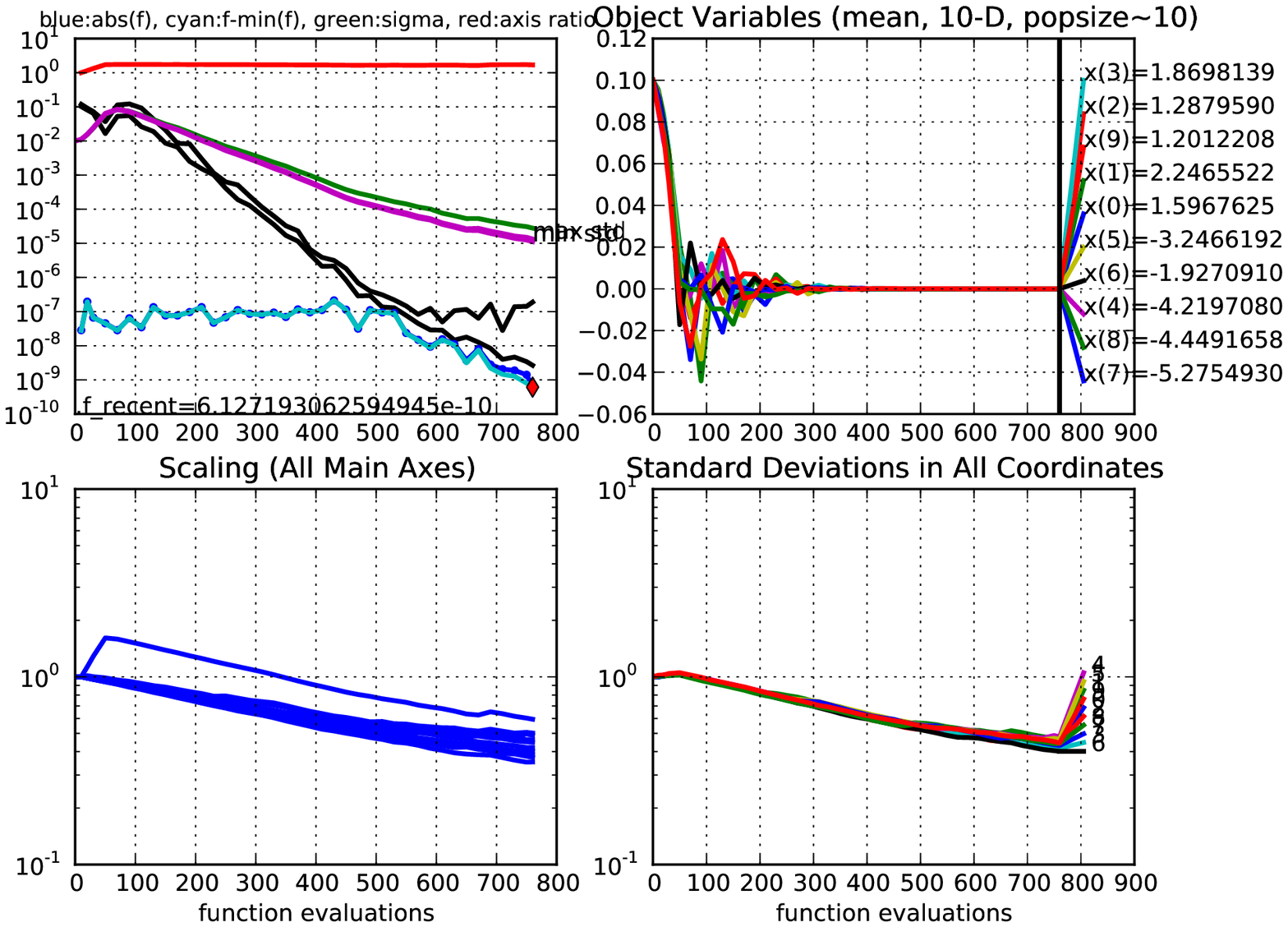}
&\includegraphics[width=0.49\textwidth, trim=0 0 11mm 8mm, clip]{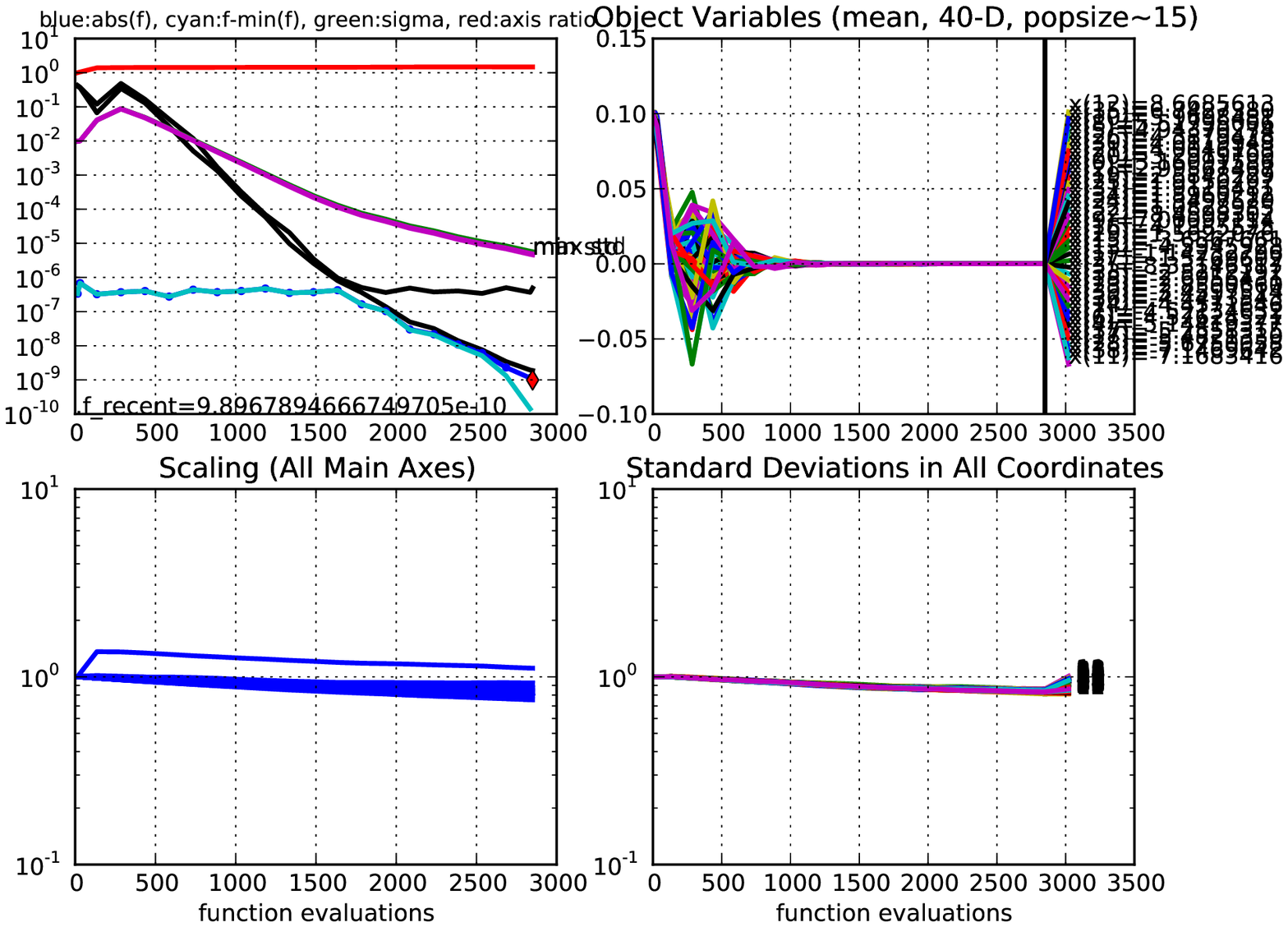}
\\\hline
\includegraphics[width=0.49\textwidth, trim=0 0 11mm 8mm, clip]{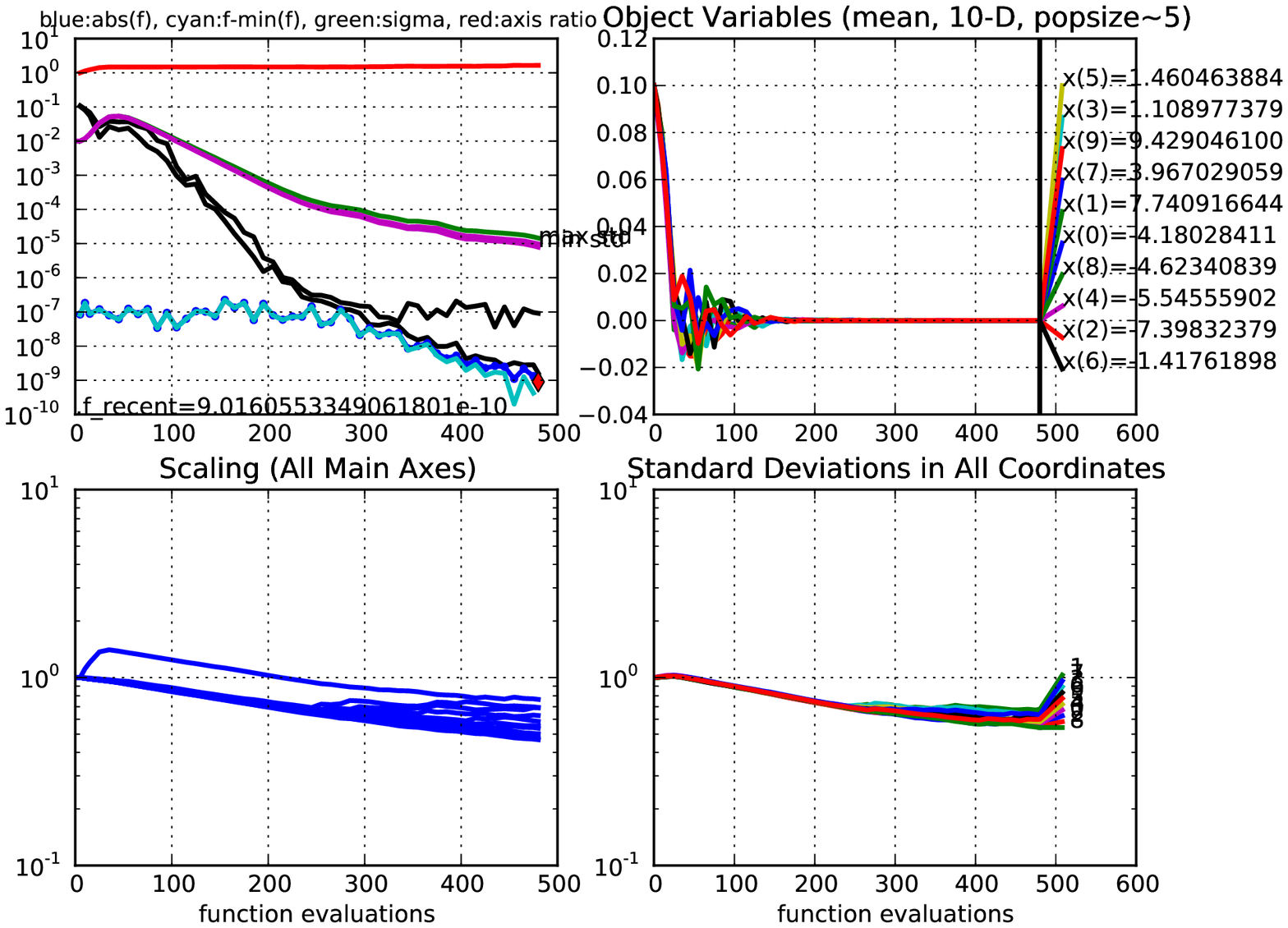}
&\includegraphics[width=0.49\textwidth, trim=0 0 11mm 8mm, clip]{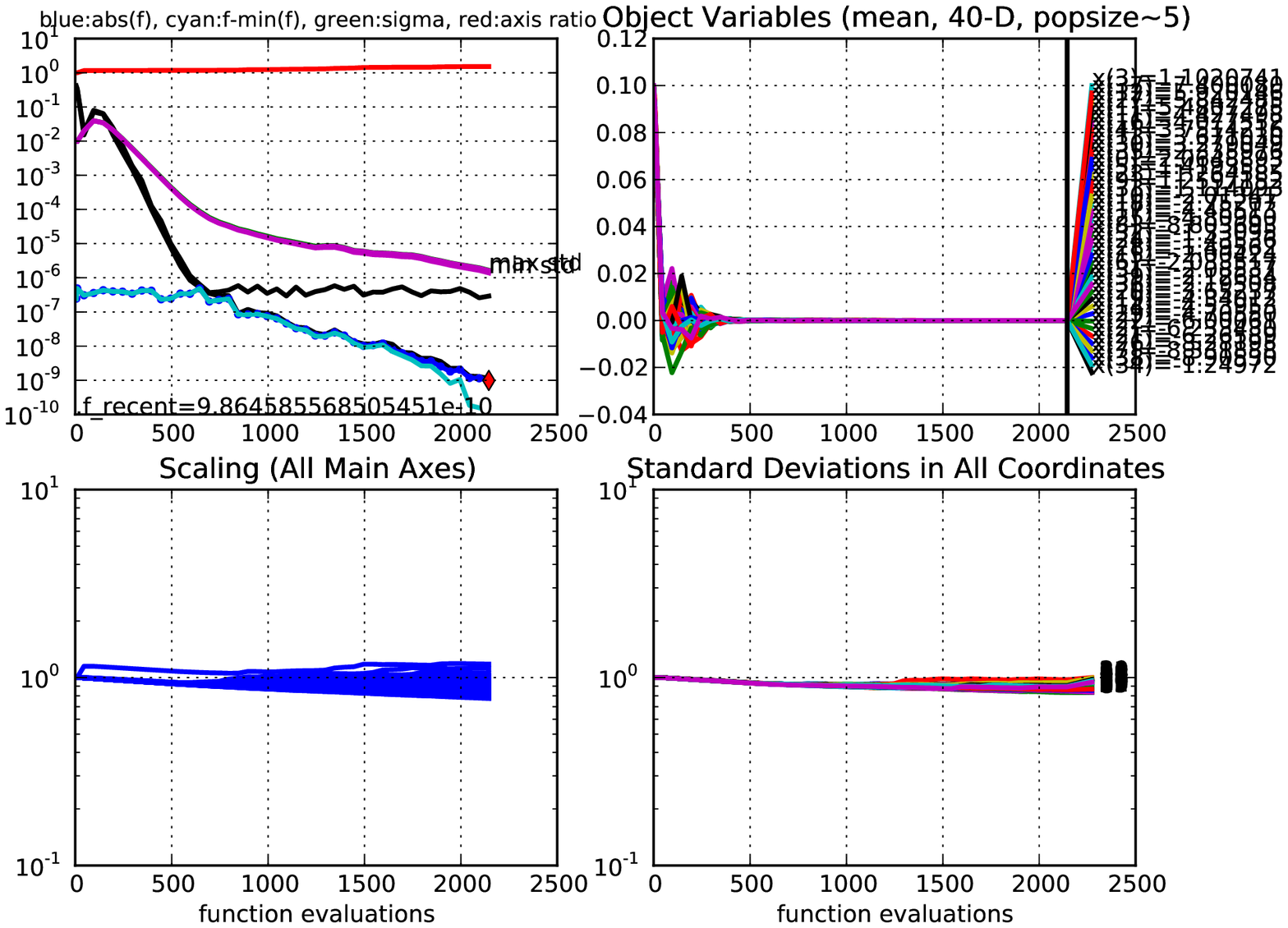}
\end{tabular}
\caption[sphere]{\label{fig:singlesphere}Runs of CMA-ES on the sphere function, middle and lower row with a single injected solution distributed as $10^{-4}\times\NormalNI$ (which corresponds in the beginning to a slightly disturbed gradient direction). Black lines show the evolution of median and worst solution. The evolution of the worst solution indicates that---with default parameter setting on the sphere function---a speed-up by a factor of two can be achieved. The reason for the comparatively moderate speed-up is that the step-size decrease per iteration is limited. With reduced population size (lower row) the speed-up increases because the number \emph{of iterations} to reach function value $10^{-6}$ in the best case scenario remains almost constant.}
\end{figure}

\begin{figure}
\begin{tabular}{@{\hspace{-3mm}}c@{}c}
 10-D & 40-D
\\
\includegraphics[width=0.49\textwidth, trim=0 0 11mm 8mm, clip]{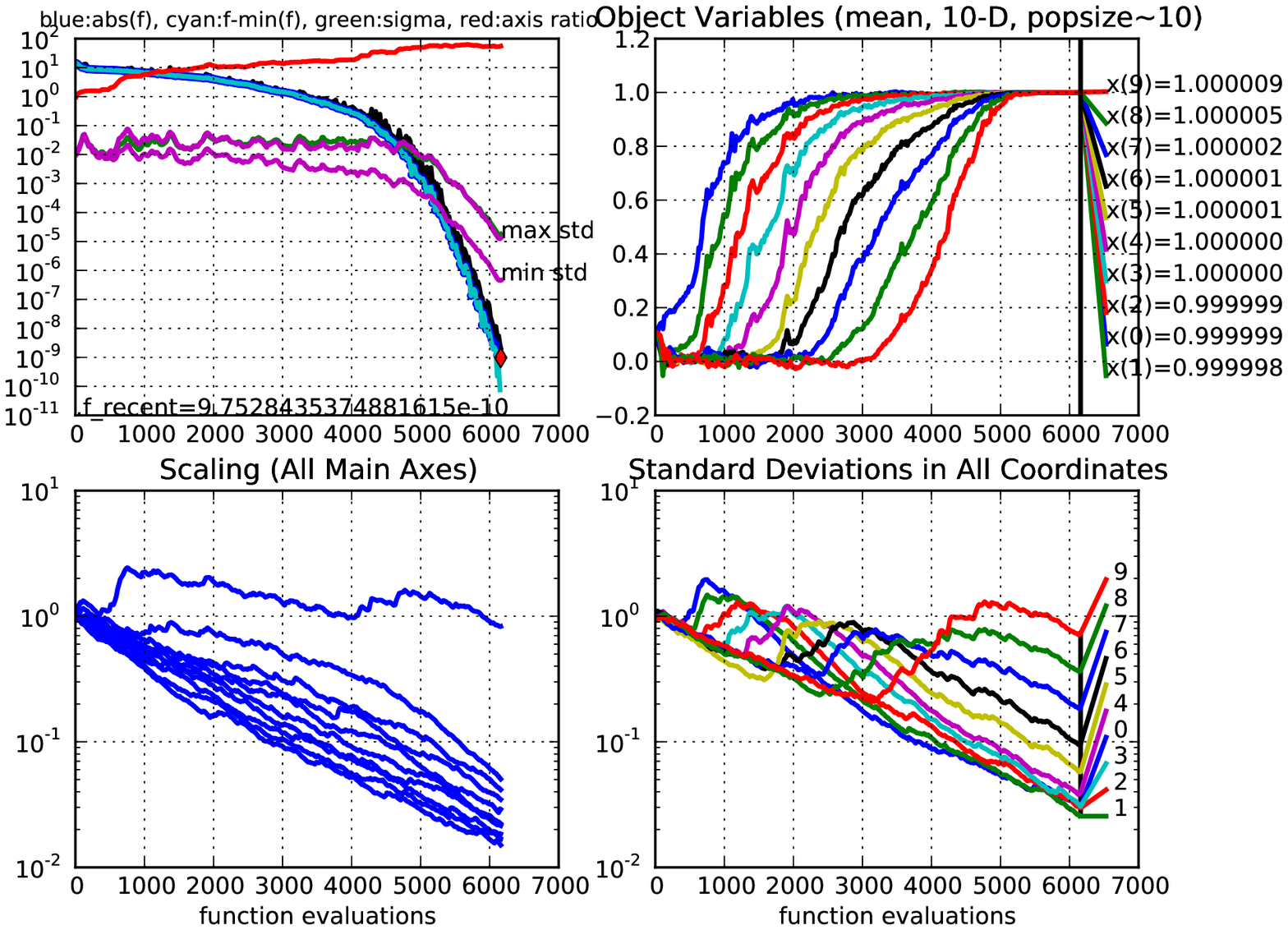}
&\includegraphics[width=0.49\textwidth, trim=0 0 11mm 8mm, clip]{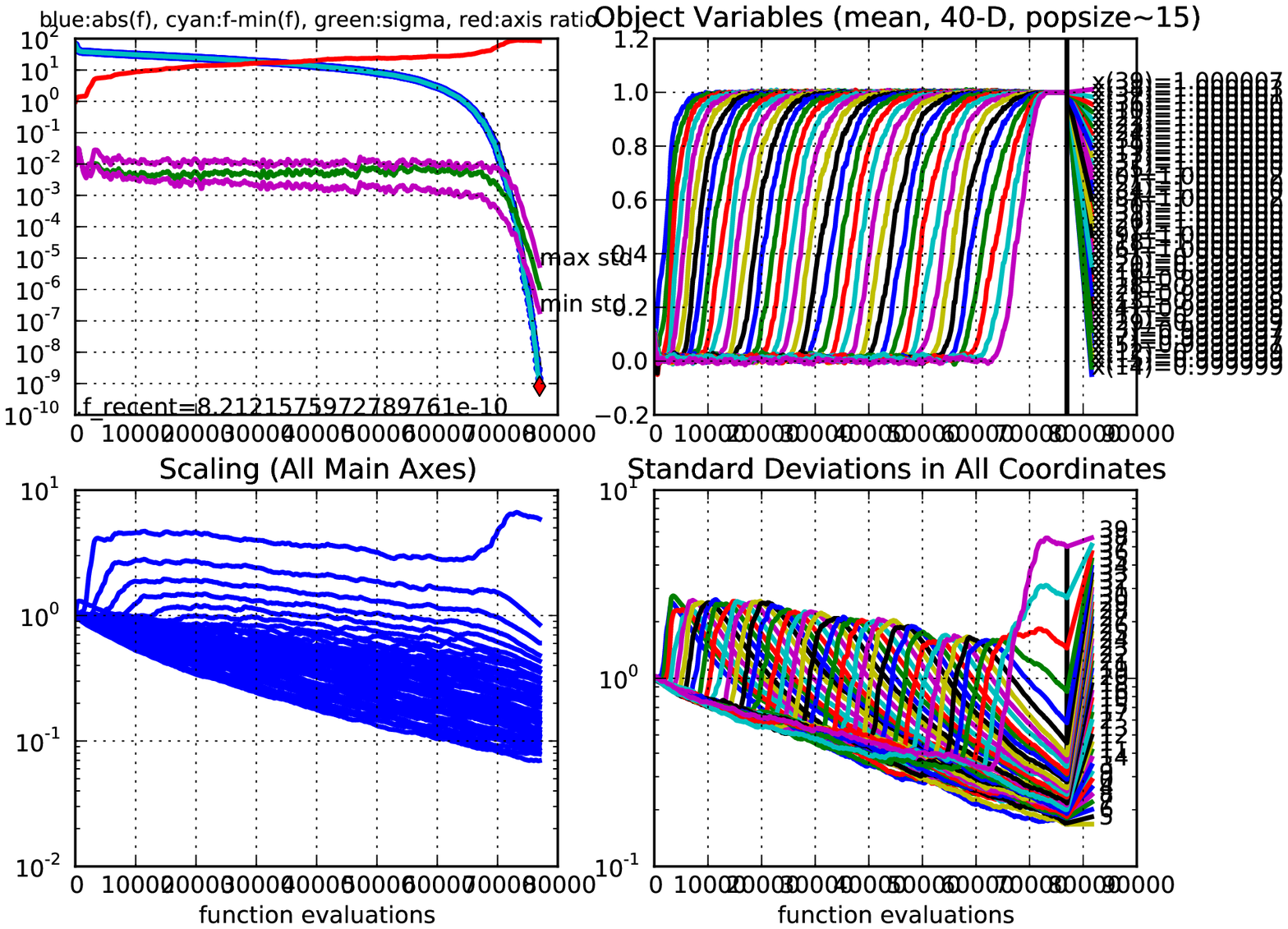}
\\\hline
\includegraphics[width=0.49\textwidth, trim=0 0 11mm 8mm, clip]{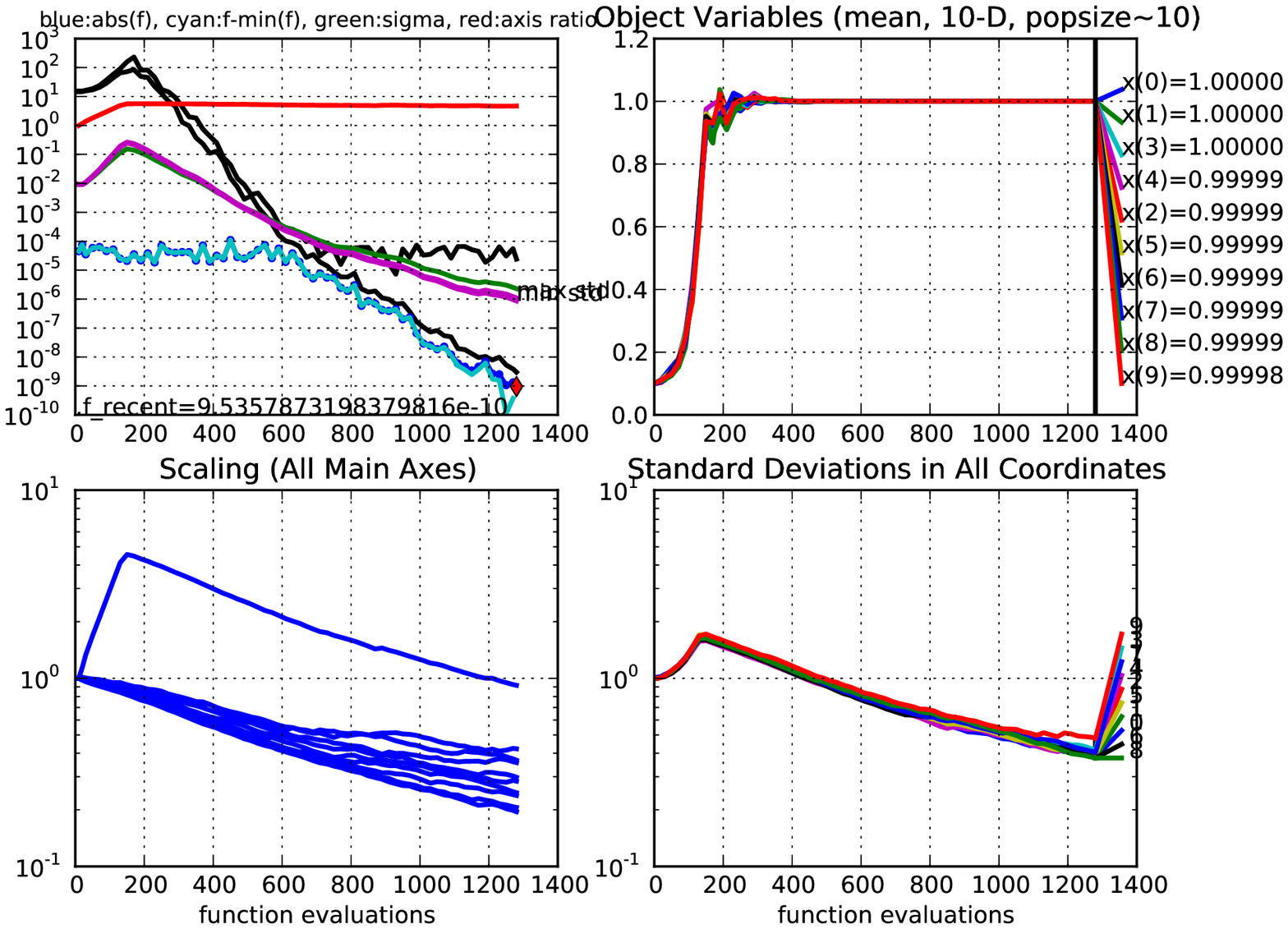}
&\includegraphics[width=0.49\textwidth, trim=0 0 11mm 8mm, clip]{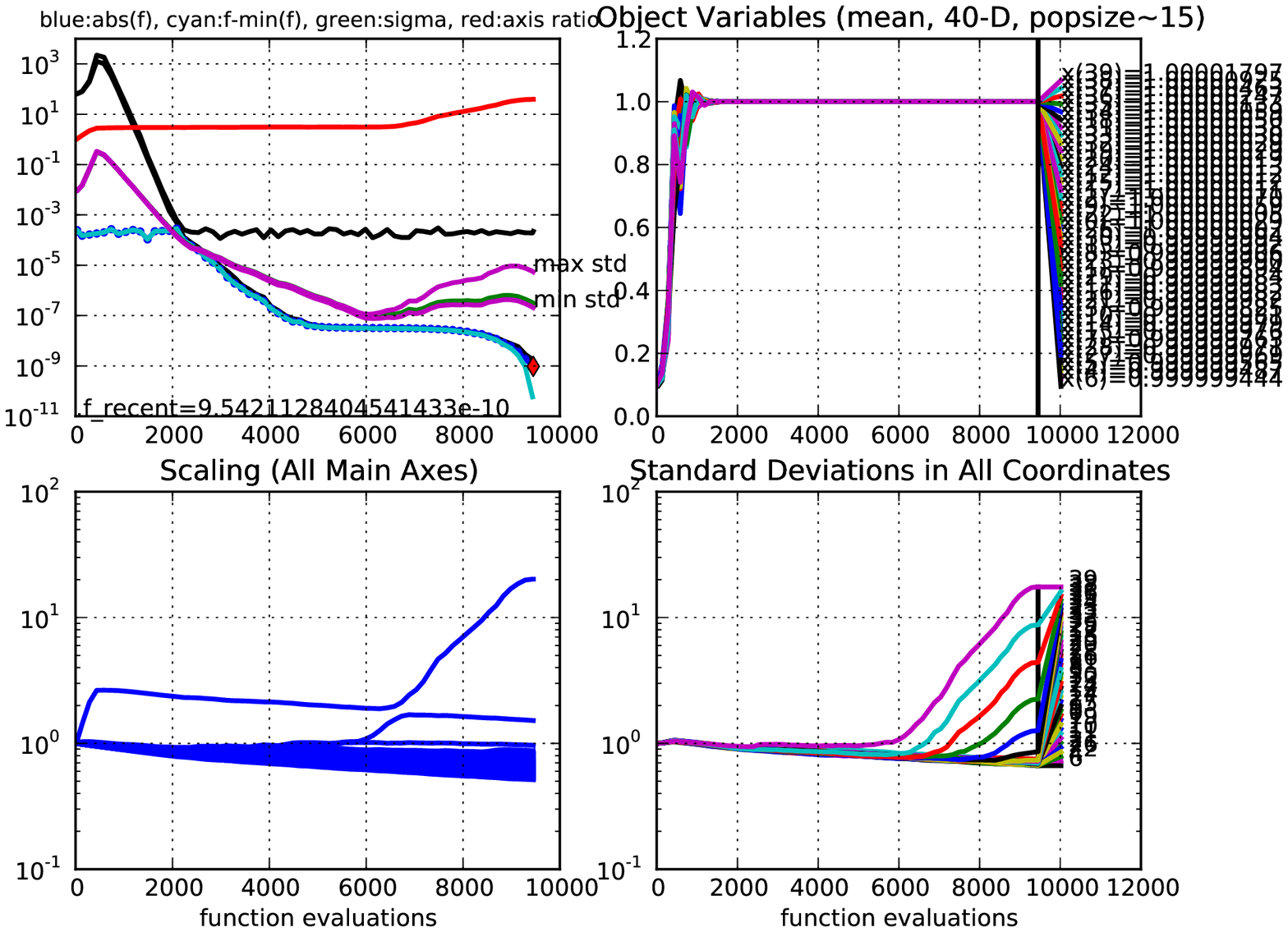}
\\\hline
\includegraphics[width=0.49\textwidth, trim=0 0 11mm 8mm, clip]{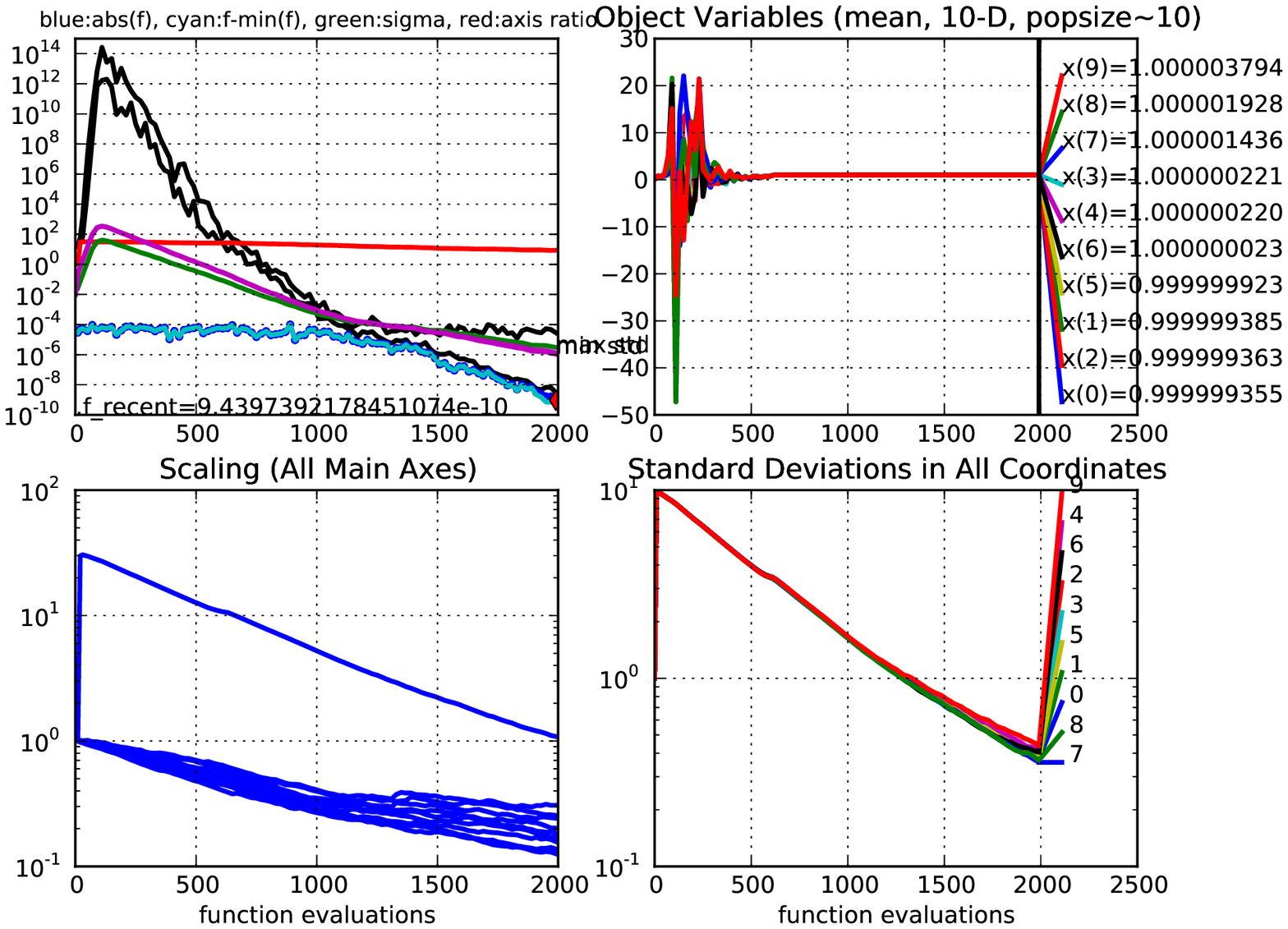}
&\includegraphics[width=0.49\textwidth, trim=0 0 11mm 8mm, clip]{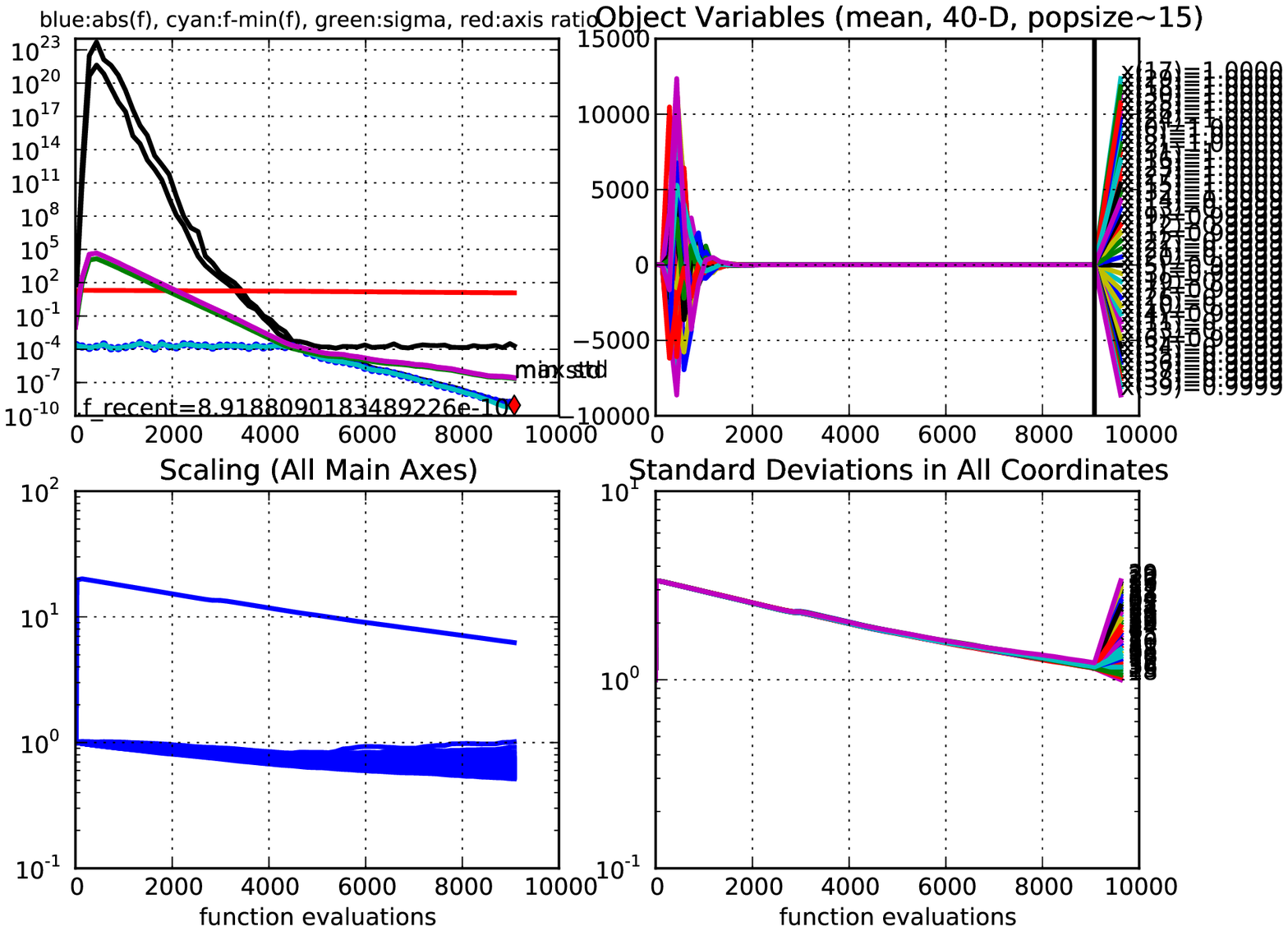}
\end{tabular}
\caption[Rosen]{\label{fig:singlerosen}Runs of CMA-ES on the Rosenbrock function, middle and lower row with a single injected solution distributed as $\ve1 + 10^{-4}\times\NormalNI$, lower row without the clipping in \eqref{eq:normalizey}. With injection, the convergence speed is again twice as fast as on the sphere function. Where the injected solutions are useful (for function values larger than about $10^{-4}$), the algorithm is almost $\dim$ times faster than without injection (600 vs 5000 and 2000 vs 70000 evaluations in 10- and 40-D). Without clipping, the run does not fail only because the step-size increment is limited to $\exp(\Dsigmax)=2.718\dots$ per iteration.}
\end{figure}

Preliminary empirical investigations have been conducted by injecting a single slightly perturbed \emph{optimal} solution. This is virtually the best case scenario when the distribution mean \m\ is far away from the optimum. This becomes the worst case scenario, when \m\ is closer to the optimum than the perturbation. Single runs on the sphere function and the Rosenbrock function are shown in Figures \ref{fig:singlesphere} and \ref{fig:singlerosen}.% 
The upper black graph depicts the worst iteration-wise function value and reveals the (sharp) transition between best and worst case scenario by showing convergence first and stagnation afterwards.

The improvement on the sphere function is limited to a factor of about two, namely due to the maximal iteration-wise step-size decrement. This limit can be exceeded by additionally decreasing $\sig$ when the injected solution is trustworthy, has a good quality, and is close to \m\ (in the norm defined by $\sig^2\C$). The precise implementation (the question of what is close to \m) might also depend on \mueff. As to be expected, the effect of injecting single bad solutions (worst case scenario in the later stage) is negligible. 

The improvement on the Rosenbrock function exceeds our expectation: we see a speed-up by a factor of almost \dim, simply because the speed is similar to the one on the sphere function with injection. Again, this speed-up can be further enhanced by step-size decrements. 

Experiments for an injected mean-shift have not been conducted yet.  

Experiments injecting always the best-ever solution reveal a moderate performance impairment when searching multimodal landscapes.

\section{Further Considerations}
Another case of application is temporary freezing of some variables (coordinates) to the same value in all candidate solutions $\x_i$. (This decreases the length of the step in the Euclidean norm, but due to correlations in the distribution this can lead to exceptionally long steps in Mahalanobis distance even if the frozen value is borrowed from \m). In this case, it is also advisable to slightly modify the step-size equations \eqref{eq:alghsig} and \eqref{eq:algsig}. Given $j$ variables are frozen, these variables are not taken into account for computing $\|\pstt\|$ and consequently $\dim-j$ is used instead of $\dim$ in \eqref{eq:alghsig} and $\chiN$ is computed for $\dim-j$ dimensions in \eqref{eq:algsig}. After one iteration, the respective components of $\y_i$ will be zero (given $\cm=1$) and also \cy\ should be set as for dimension $\dim-j$. In principle, all parameters from Table~\ref{tabdef} can then be set as for dimension $\dim-j$. Additionally, in order to avoid numerical problems, the diagonal elements of the frozen coordinates of the covariance matrix should be kept at least in the order of the smallest eigenvalue.  

\section{Summary and Conclusion}
Using candidate proposals in the CMA-ES that do not directly stem from the sample distribution of CMA-ES can often lead to a failure of the algorithm.  The effective counter measures however turn out to be comparatively simple: only the appearance of large steps needs to be tightly controlled, where large is defined w.r.t.\ the original sample distribution. The possibility to inject any candidate solution is valuable in many situations. In case of bounds or constraints where a repair mechanism is available, this might serve as basis for a new class of well-performing constraint handling mechanisms.   

\paragraph{Acknowledgment}
This work was supported by the ANR-2010-COSI-002 grant (SIMINOLE) of the French National
Research Agency.

\bibliographystyle{plain}
% \bibliography{optimbib}

\end{document}